\documentclass[10pt, conference]{format/ieeeconf}

\IEEEoverridecommandlockouts    
\usepackage{glossaries}
\newacronym{mav}{MAV}{Micro Aerial Vehicles}
\newacronym{uav}{UAV}{Unmanned Aerial Vehicle}
\newacronym{ovc}{OVC}{Open Vision Computer}
\newacronym{lidar}{LiDAR}{Light Detection and Ranging}
\newacronym{vio}{VIO}{visual-inertial odometry}
\newacronym{gpgpu}{GPGPU}{General-Purpose Graphics Processing Unit}
\newacronym{ugv}{UGV}{Unmanned Ground Vehicle}
\newacronym{uwb}{UWB}{Ultra Wideband}
\newacronym{svm}{SVM}{Support Vector Machine}
\newacronym{fcn}{FCN}{Fully Convolutional Network}
\newacronym{cnn}{CNN}{Convolutional Neural Network}
\newacronym{loam}{LOAM}{LiDAR Odometry and Mapping}
\newacronym{sloam}{SLOAM}{Semantic LiDAR Odometry and Mapping}
\newacronym{slam}{SLAM}{Simultaneous Localization and Mapping}
\newacronym{iot4ag}{IoT4Ag}{NSF Engineering Research Center for the Internet of Things for Precision Agriculture}
\newacronym{grasp-lab}{GRASP Lab}{the General Robotics, Automation, Sensing and Perception Laboratory}
\newacronym{jps}{JPS}{Jump Point Search}
\newacronym{ukf}{UKF}{Unscented Kalman Filter}
\newacronym{sam}{SAM}{Smoothing and Mapping}
\newacronym{icp}{ICP}{Iterative Closest Point}
\newacronym{imu}{IMU}{Inertial Measurement Unit}
\newacronym{tsdf}{TSDF}{Truncated Signed Distance Field}
\newacronym{esdf}{ESDF}{Euclidean Signed Distance Field}
\newacronym{sdf}{SDF}{Signed Distance Field}
\newacronym{rrt}{RRT}{Rapidly Exploring Random Tree}
\newacronym{fpv}{FPV}{First-person View}
\newacronym{dnn}{DNN}{Deep Neural Network}
\newacronym{igpred}{IGPred}{Information Gain Prediction}
\newacronym{csqmi}{CSQMI}{Cauchy-Schwarz Quadratic Mutual Information}
\newacronym{nbv}{NBV}{Next Best View}
\newacronym{vae}{VAE}{Variational Autoencoder}
\newacronym{tsp}{TSP}{Travelling Salesman Problem}
\newacronym{bgsm}{BGSM}{Behavior Guidance State Machine}
\newacronym{pca}{PCA}{Principal Component Analysis}
\newacronym{aspp}{ASPP}{Atrous Spatial Pyramid Pooling}
\newacronym{swap}{SWaP}{Size Weight and Power}
\newacronym{soi}{SoI}{Semantic Object of Interest}
\newacronym{aoi}{AoI}{Area of Interest}

\usepackage{tabularx}
\usepackage{multirow, multicol}
\usepackage{array}
\newcolumntype{P}[1]{>{\centering\arraybackslash}p{#1}}
\newcolumntype{M}[1]{>{\centering\arraybackslash}m{#1}}
\usepackage{booktabs}
\newcolumntype{N}{>{\centering\arraybackslash}m{.5in}}
\newcolumntype{G}{>{\centering\arraybackslash}m{2in}}

\newcommand\edited[1]{\textcolor{black}{#1}}

\let\labelindent\relax

\usepackage[table,usenames,dvipsnames]{xcolor}      
\usepackage{extarrows}                              
\usepackage{enumitem}
\usepackage{wrapfig}

\usepackage{amsmath,amssymb,amsfonts,amsthm,dsfont, bm} 
\usepackage{mathtools}
\usepackage{algorithm,algorithmicx,listings}        
\usepackage[noend]{algpseudocode}			              
\makeatletter
\def\BState{\State\hskip-\ALG@thistlm}
\makeatother
\DeclarePairedDelimiter\abs{\lvert}{\rvert}%
\DeclarePairedDelimiter\norm{\lVert}{\rVert}%
\makeatletter
\let\oldabs\abs
\def\abs{\@ifstar{\oldabs}{\oldabs*}}
\let\oldnorm\norm
\def\norm{\@ifstar{\oldnorm}{\oldnorm*}}
\makeatother

\usepackage{graphicx}
\usepackage{subcaption}
\usepackage{textcomp}
\usepackage[font=footnotesize]{caption}
\usepackage{subcaption}
\usepackage[breaklinks=true, colorlinks, bookmarks=true, citecolor=Black, urlcolor=Violet,linkcolor=Black]{hyperref}

\usepackage[normalem]{ulem}
\usepackage[utf8]{inputenc}
\usepackage[english]{babel}
\usepackage{hyperref}
\hypersetup{
    colorlinks=true,
    linkcolor=blue,
    filecolor=magenta,      
    urlcolor=cyan,
}

\DeclareMathAlphabet\mathbfcal{OMS}{cmsy}{b}{n}

\usepackage[space, compress, sort]{cite}

\newtheorem*{assumption*}{Assumption}

\newtheorem*{problem*}{Problem}

\usepackage{lipsum}
\usepackage[capitalise]{cleveref}

\usepackage{tikz}

\newcommand\copyrighttext{%
  \footnotesize \textcopyright 2023 IEEE. Personal use of this material is permitted.
  Permission from IEEE must be obtained for all other uses, in any current or future
  media, including reprinting/republishing this material for advertising or promotional
  purposes, creating new collective works, for resale or redistribution to servers or
  lists, or reuse of any copyrighted component of this work in other works.}
\newcommand\copyrightnotice{%
\begin{tikzpicture}[remember picture,overlay]
\node[anchor=south,yshift=10pt] at (current page.south) {\fbox{\parbox{\dimexpr\textwidth-\fboxsep-\fboxrule\relax}{\copyrighttext}}};
\end{tikzpicture}%
}

\begin{document}

\title{SEER: Safe Efficient Exploration for Aerial Robots\\ using Learning to Predict Information Gain}

\author{Yuezhan Tao, Yuwei Wu, Beiming Li, Fernando Cladera, Alex Zhou, Dinesh Thakur and Vijay Kumar 
\thanks{All authors are with GRASP Laboratory, University of Pennsylvania {\tt\small\{yztao, yuweiwu, beimingl, fclad, alexzhou, tdinesh, kumar\}@seas.upenn.edu}.} %
\thanks{We gratefully acknowledge the support of
ARL DCIST CRA W911NF-17-2-0181, 
NSF Grants CCR-2112665, 
ONR grant N00014-20-1-2822, 
ONR grant N00014-20-S-B001, 
NVIDIA, 
\edited{Israel Minister of Defense,}
and C-BRIC, a Semiconductor Research Corporation Joint University Microelectronics Program program cosponsored by DARPA. 
We also thank Yifei (Simon) Shao for providing helpful insights and being our safety pilot in the real-world experiments.}
\thanks{Digital Object Identifier (DOI): 10.1109/ICRA48891.2023.10160295}
}
\maketitle
\copyrightnotice

\begin{abstract}

We address the problem of efficient 3-D exploration in indoor environments for micro aerial vehicles with limited sensing capabilities and payload/power constraints. We develop an indoor exploration framework that uses learning to predict the occupancy of unseen areas, extracts semantic features, samples viewpoints to predict information gains for different exploration goals, and plans informative trajectories to enable safe and smart exploration. Extensive experimentation in simulated and real-world environments shows the proposed approach outperforms the state-of-the-art exploration framework by $\textbf{24\%}$ in terms of the total path length in a structured indoor environment and with a higher success rate during exploration.

\end{abstract}
    \section{Introduction}
\label{sec:intro}

Autonomous exploration requires a robot to navigates through an unknown region and map it in real time. This problem has always attracted great attention in the robotics community due to its direct applications in tasks such as search and rescue, precision agriculture, and autonomous inspection. All of these tasks require the robot to navigate through cluttered and complex environments. Therefore, \gls{mav}, especially quadrotors, have gained great popularity due to their agility.

Autonomous exploration has been widely studied and recent works have measured its performance in terms of map uncertainty~\cite{KelseyIG, LukasIG, charrow2015information}, total travel distance~\cite{zhou2021fuel}, energy efficiency~\cite{RapidExploration}, and exploration time~\cite{zhou2021fuel, alexis2020MP}. Beyond optimizing for different objectives, different exploration strategies and planning methods have been proposed to tackle different environments, such as structured indoor environments~\cite{topo2020indoor, shen2012indoor, 2dMapPred2019, 2d2019semanticIndoorExp, ramon2019semanticIndoor} and subterranean environments~\cite{2dtopoPred2020, alexis2020MP}.

Nevertheless, most of the existing works follow heuristics to greedily expand the map or maximize the information to mimic human behavior; they do not consider inferring the importance of the environment through semantic objects. Although some methods do consider semantic information to guide the exploration, this information is not fully utilized during local planning. Furthermore, in many works information gain is optimistically estimated without considering the environment. Most of these works assume that all space inside the robot FOV at the goal point will be observed, leading to an overestimate of the information gain. 

\begin{figure}[!t]
    \centering
    \includegraphics[width=1.0\columnwidth]{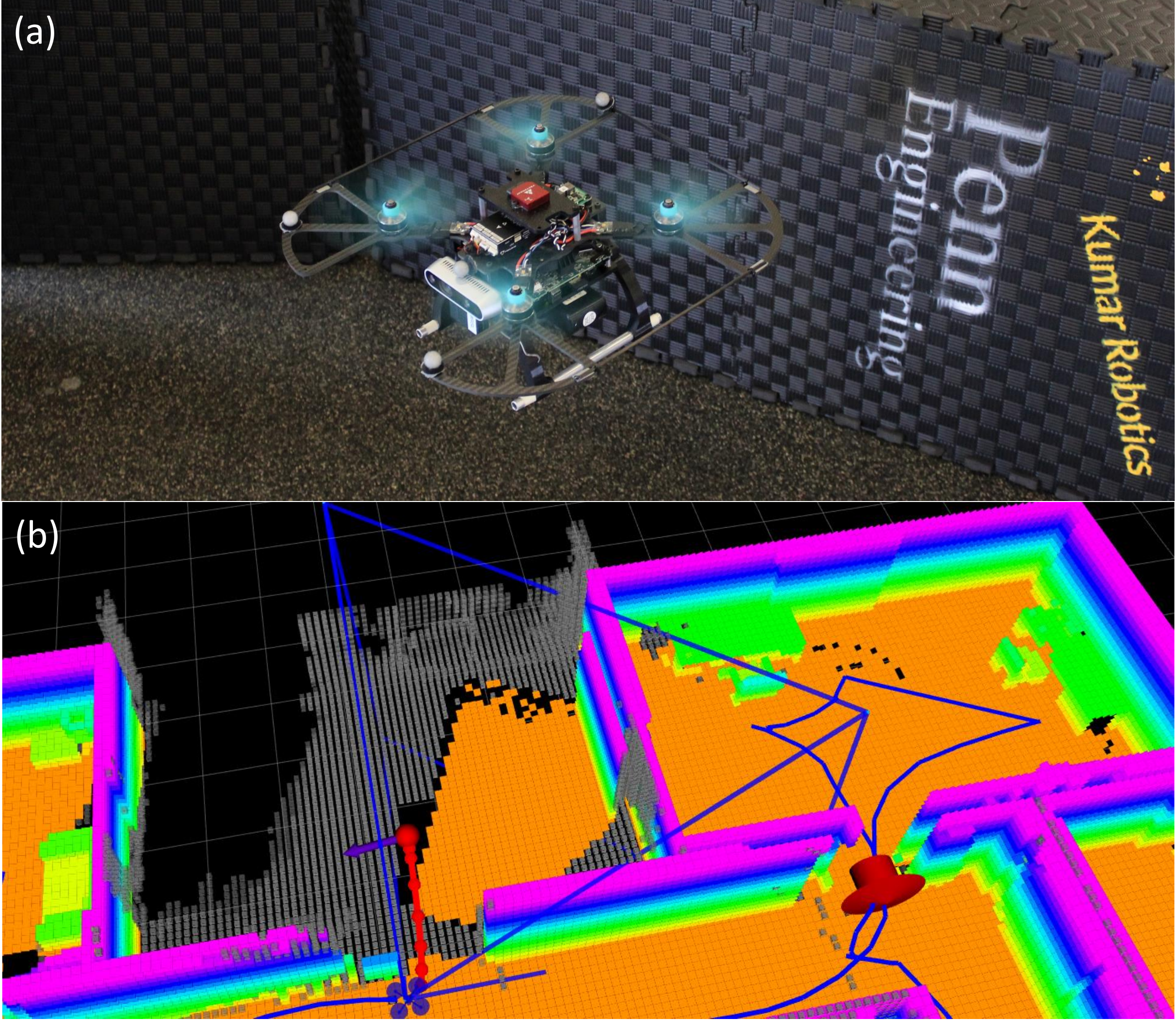}
    \caption{(a) \edited{Falcon} 250 Platform. The \edited{Falcon} 250 platform is equipped with an Intel Realsense D435i camera, an Intel NUC 10 onboard computer with an i7-10710U CPU, and a Pixhawk 4 Mini flight controller. (b) A robot exploring a simulated environment with the proposed method. The predicted map is represented by gray cubes, executed trajectories are shown as blue curves, detected semantic objects are shown as red arrows, the next viewpoint is the purple arrow, and the planned trajectory is shown in red.}
    \label{fig:fig1}
    \vspace{-.5cm}
\end{figure}

Motivated by these limitations, we propose an indoor \gls{mav} 3-D exploration framework that uses high-level behavioral policies and predicts information gain for intelligent and efficient exploration. Our work takes inspiration from the behavior of humans during exploration, i.e., inferring the importance of the environment by identifying and using semantic information. In addition, humans predict the structure of the unknown area by making assumptions based on observed regions. Therefore, in our proposed framework, we propose a detection and prediction module that detects frontier clusters, extracts semantic information, and predicts occupancy and information incrementally during exploration. 
This abstracted information is then used by a perception-aware planning module in which the behavior planning sets the global navigation goal following the behavior policies and the predicted informative planning generates local trajectories that allow safe and informative maneuvers. An illustration of our proposed system is shown in Fig.~\ref{fig:fig1}.

We benchmarked our proposed framework with classical and state-of-the-art methods in simulation experiments. We also conducted real-world experiments in which our proposed framework runs fully on board a \gls{swap} constrained platform in a structured indoor environment. In summary, \textbf{our contributions are}:
\begin{itemize}
    \item An incremental detection and prediction module that detects semantic objects, classifies frontiers, predicts information gain, and samples viewpoints around clustered frontiers.
    \item A perception-aware planning approach that guides the exploration through behavior planning and plans predicted informative trajectories to ensure intelligent, safe, and fast exploration in a structured indoor environment.
    \item We validate the proposed framework and algorithms extensively in simulation and real-world experiments. We will open-source our proposed system and algorithm at \url{https://github.com/tyuezhan/SEER}.
\end{itemize}

    \section{Related Work}
\label{sec:related work}

\subsection{Autonomous Exploration Strategy}
Various exploration strategies have been proposed to autonomously map unknown regions. There are three
major exploration strategies: frontier-based, information-based, and the hierarchical mixed approach.

Following the idea of purely expanding the map into unknown regions, frontier-based exploration~\cite{Frontier} has been widely adopted. Real-world demonstrations in \cite{frontier2012_system, shen2012indoor} show its applicability in both 2-D and 3-D environments. To achieve high energy efficiency during exploration, \cite{RapidExploration} proposes to select the frontier within the agent's FOV to maintain a high flight speed. Instead of searching frontiers, maximizing the information gain serves as another heuristic to guide exploration. \gls{nbv} is sampled from \gls{rrt}~\cite{NBVexplore2016, bircher2018receding} or motion primitives~\cite{alexis2020MP} to maximize information gain. Different methods are also proposed to evaluate the information gain~\cite{charrow2015CSQMI, KelseyIG, LukasIG}. However, classical methods assume all voxels in the ray-casting process will be updated, resulting in an overestimate of total information gain. Hierarchical approaches have inherent advantages by benefiting from multiple strategies. In~\cite{charrow2015information, mixed2019SampleVP}, global paths are planned towards frontiers while local paths are sampled through motion primitives and optimized for information gain. \cite{mixed2017FrontierTSP, zhou2021fuel} plans a global coverage path through \gls{tsp} to efficiently travel through all viewpoints and sample viewpoints around frontiers to maximize coverage or information gain. To further improve the efficiency and consistency of exploration, semantic and topological information is used in~\cite{topo2020indoor, ramon2019semanticIndoor} to assist high-level robot behaviors. However, directly incorporating the semantic information into the utility function sometimes fails to guide the exploration, as the semantic information might be labelled incorrectly. In~\cite{2d2019semanticIndoorExp}, \gls{nbv} is selected following a decision model to guide the exploration behavior. However, it assumes that a semantic mapping model will provide perfect semantic information of the environment. 

In this paper, we present a perception-aware planning framework in which the behavior planning module takes the detected semantic information into consideration to select a candidate goal, and the predicted informative planning module utilizes the predicted information gain to improve exploration efficiency.

\subsection{Map Completion and Prediction}
Predicting the occupancy of unknown environments can provide prior information to help with navigation and exploration tasks. Database-based approaches and deep learning-based approaches are commonly used. To infer unknown information, historical data is stored and used in~\cite{mapRetrival2015, smith2018distributedInference}. Without using a database, \cite{pimentel20182dinformationExtrapolation} extrapolates the local 2D structure to estimate the potential information gain. With the capability of learning from and adapting to a large amount of historical data, map prediction using deep learning has gained popularity in recent years. 2D occupancy maps are predicted in~\cite{2dMapPred2019, uncertaintyPrediction2019} to estimate the total information or uncertainty for exploration. In~\cite{2dtopoPred2020}, topological information is predicted in a large-scale underground environment. Egocentric 2D occupancy and semantic maps are predicted in~\cite{OccUnvertaintyPrediction2022, https://doi.org/10.48550/arxiv.2106.15648} to help the navigation tasks. To achieve safe and fast navigation, \cite{3DoccPred2021-OPNet} proposed egocentric 3-D occupancy prediction of occluded areas. Most existing work focuses on using 2D structural information to predict the occupancy and estimate the total potential information gain within a bounding box. This approach provides a good estimate of the total information that can be obtained in a large region. However, it overestimates the information, as it cannot infer unseen environment. It also does not help with estimating the information gain at certain viewpoints in a 3-D environment. 

In this paper, utilizing 3-D occupancy prediction, we present a method to predict information gain in unknown regions beyond frontiers to sample the viewpoint which maximizes the potential information gain accurately.

    \section{System Overview}
\label{sec:system_overview}

\begin{figure}[!t]
    \centering
    \includegraphics[width=1.0\columnwidth]{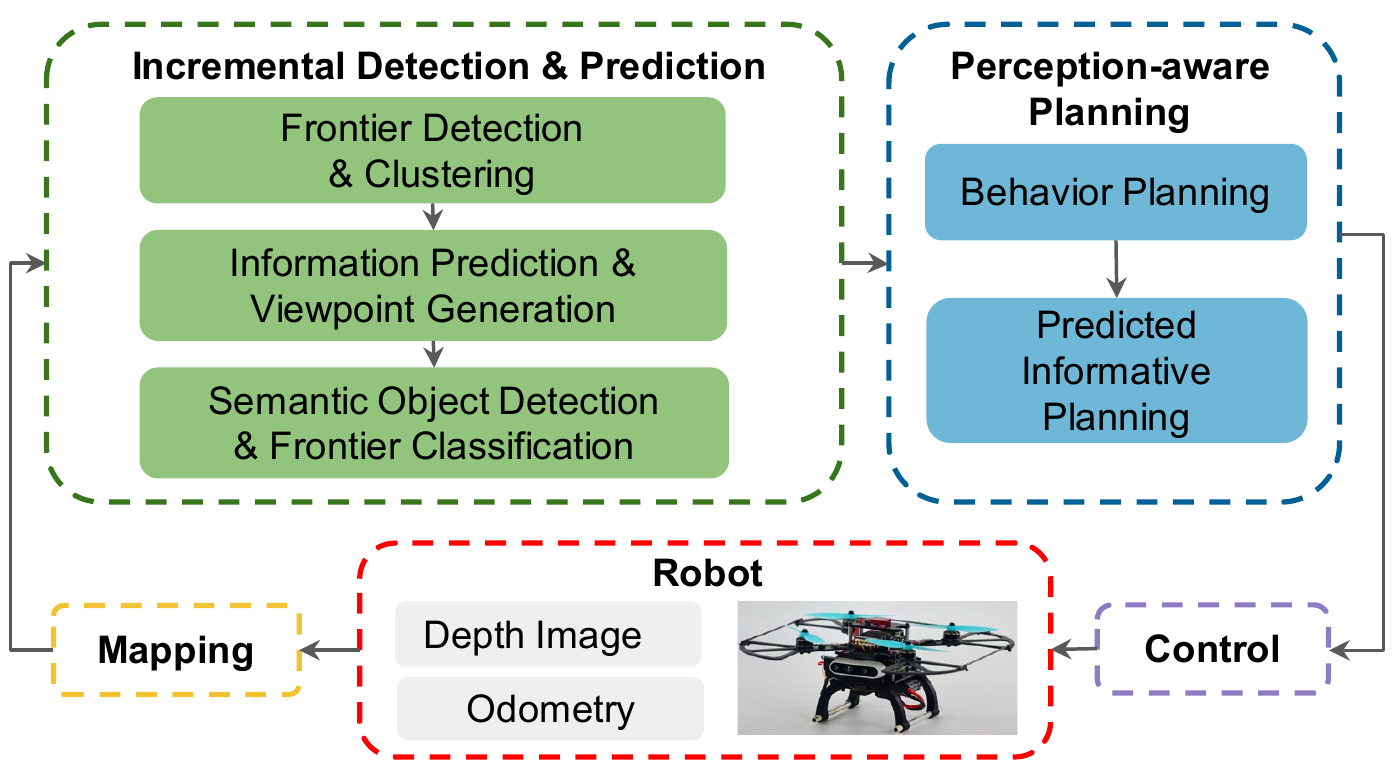}
    \caption{Proposed system architecture. The proposed system contains two main components: the incremental detection \& prediction module, and the perception-aware planning module.}
    \label{fig:sys_diagram}
    \vspace{-.5cm}
\end{figure}

As illustrated in Fig.~\ref{fig:sys_diagram}, the proposed system contains two main components.

The incremental detection and prediction module (Sect.~\ref{sec:detection_and_prediction}) first takes in a bounding box of the updated map, reevaluates existing frontiers, detects new frontiers, and clusters them. Then, semantic object detection, frontier classification, and occupancy prediction are executed around clusters. Viewpoints are sampled and sorted according to the predicted information gain. The perception-aware planning module (Sect.~\ref{sec:hybrid_planning}) then uses this information to generate a two-stage exploration plan that enables safe and informative flight.

    \section{Incremental Detection and Prediction}
\label{sec:detection_and_prediction}
During the exploration phase, the map of the environment is built incrementally, so it is natural to evaluate existing frontiers and detect new frontiers from the updated regions of the map. Similarly, we incrementally detect semantic objects, label frontier clusters, and sample informative viewpoints through accurate information prediction.

\subsection{Frontier Detection and Clustering}
\label{sec:detection_and_prediction_clustering}
We implemented the frontier detection and clustering algorithm of~\cite{zhou2021fuel}.
The incremental frontier detection and clustering module receives a bounding box that represents the updated region of the map. Existing frontiers inside the bounding box are reevaluated, and frontiers that have been changed are removed. New frontiers are searched and clustered inside the bounding box by a region-growing algorithm, and large clusters are split into smaller ones recursively by \gls{pca} along the principal axis if the largest eigenvalue exceeds a predefined threshold. Each time a new cluster $F_{i}$ is detected, a frontier cluster node is created and saved.

\subsection{Information Prediction \& Viewpoint Generation}
\label{sec:detection_and_prediction_info_pred}
As frontiers are the boundary between the known and unknown regions, it is natural to navigate towards them to expand the map into the unknown regions. To find a goal that provides the most information from the unknown environment, we present a novel, combined information prediction and viewpoint generation method that takes the occupancy prediction into account to predict future information gain accurately.

The occupancy grid map is represented as a vector $\mathbf{M}$ of size $N = {N_x \cdot N_y \cdot N_z}$ where $N_x, N_y, N_z$ are the sizes of the grids along each axis. Each element $m_{xyz} \in (clamp_{min}, clamp_{max})$ represents the log-odds of the probability of the occupancy for the voxel. For occupancy prediction, we cut-off the occupancy values $m_{xyz}$ into trinary ${o}_{xyz} \in \{-1, 0, 1\}$ to generate $\mathbf{O}$, where $\{-1, 0, 1\}$ denotes $\{unknown, free, occupied\}$, respectively.

\subsubsection{3-D Occupancy Prediction}
We generate realistic training data pairs to simulate the partially observed scenes during exploration for the self-supervised 3-D occupancy prediction network. After clustering the frontiers, we extract the local region centered at the centriod of each frontier cluster $F_{i}$ from the volumetric map and perform occupancy prediction for unknown regions beyond the frontiers. Taking into account the gap between simulation environments and real-world scenarios, we choose the MatterPort3D dataset~\cite{Matterport3D}, which contains depth images and 6-DoF camera poses collected from 90 real-world buildings. To generate the target occupancy data $\mathbf{O}^{target}$, we first reconstruct the 3-D occupancy grid map of the buildings from the raw depth images with $10cm$ resolution. Then we crop a $8m \times 8m \times 2.4m$ region at the center of each scene. Since some of the buildings contain multiple layers, we align the ground to the bottom of the cropped region to ensure consistency of the data. To generate the corresponding input data $\mathbf{O}^{in}$, we first subsample $50\%$ of the input depth images at each scene during map reconstruction. To make $\mathbf{O}^{in}$ more similar to the exploration scenarios, we further crop it with straight lines with different slopes through the center of the box. With this operation, we also increased the number of training data pairs. As a result, we generate $93,768$ $(\mathbf{O}^{in}, \mathbf{O}^{tar})$ pairs from 78 buildings in MatterPort3D.

To predict $\mathbf{O}^{tar}_{known} \setminus \mathbf{O}^{in}_{unknown}$, the known occupancy value in the target that is unknown in the input, we use OPNet~\cite{3DoccPred2021-OPNet} which contains a U-Net~\cite{unet} with \gls{aspp}~\cite{ASPP} to improve the prediction on a multiscale level. As the goal of the occupancy prediction is to provide occupancy information beyond the frontiers for the estimation of information gains, we design a loss function $\mathcal{L}$ that contains an occupancy loss $\mathcal{L}_{occ}$ that emphasizes the prediction of the occupancy of individual voxels and a structural loss $\mathcal{L}_{struct}$ that focuses on the inference of unknown structures. The loss is defined as follows:
\begin{equation}
    \mathcal{L} = \omega_{occ} * \mathcal{L}_{occ} + \omega_{struct} * \mathcal{L}_{struct}
\end{equation}
where $\omega_{occ}$ and $\omega_{struct}$ are hyperparameters.  
For occupancy loss $\mathcal{L}_{occ}$, we compute weighted binary cross-entropy (BCE) between the prediction $\mathbf{O}^{pred}$ and the target $\mathbf{O}^{tar}$:
\begin{equation}
    \begin{split}
       \mathcal{L}_{occ} = & -\frac{1}{N} \sum_{x=1}^{N_x} \sum_{y=1}^{N_y} \sum_{z=1}^{N_z} \lambda_{xyz} [  o_{xyz}^{tar} \cdot \log o_{xyz}^{pred} \\
           & + (1-o_{xyz}^{tar}) \cdot \log (1-o_{xyz}^{pred}) ] \\
              \lambda_{xyz} & =
    \begin{cases}
      0 & \text{if $o_{xyz}^{tar} = -1$} \\
      \alpha & \text{if $o_{xyz}^{in} = -1$ and $o_{xyz}^{tar} = 1 $}\\
      1 & \text{otherwise}
    \end{cases}
    \end{split}
\end{equation}
where $\alpha$ is a weight to emphasize the prediction of occupied voxels on the target occupancy map. The structural loss considers the weighted $L1$ norm between the number of occupied voxels in the target along the $Z$ axis and the number of occupied voxels in the prediction along the $Z$ axis to emphasize the prediction of unseen structures. The structural loss is defined as:
\begin{equation}
    \begin{split}
    & \begin{split}
       \mathcal{L}_{struct}  = & \frac{1}{N_x \cdot N_y} \sum_{x=1}^{N_x} \sum_{y=1}^{N_y} \varphi_{xy} \left| \sum_{z=1}^{N_z} \mathds{1}_{\{o_{xyz}^{pred} = 1\}}  \right.\\
           & \left. - \sum_{z=1}^{N_z} \mathds{1}_{\{o_{xyz}^{tar} = 1\}} \right| \\
    \end{split} \\
   &  \varphi_{xy} =\begin{cases}
      \beta & \text{if $\sum_{z=1}^{N_z} \mathds{1}_{\{ o_{xyz}^{tar} \}} > N_z / 2 $  } \\
      1 & \text{otherwise}
    \end{cases}
    \end{split}
\end{equation}
where we set weight $\beta$ on the structural information along the $z$ axis. As the frontiers are clustered, the occupancy prediction is performed on each $F_{i}$ incrementally.

\subsubsection{Information Prediction and Viewpoint Generation}
\begin{figure}[!t]
    \centering
    \includegraphics[width=1.0\columnwidth]{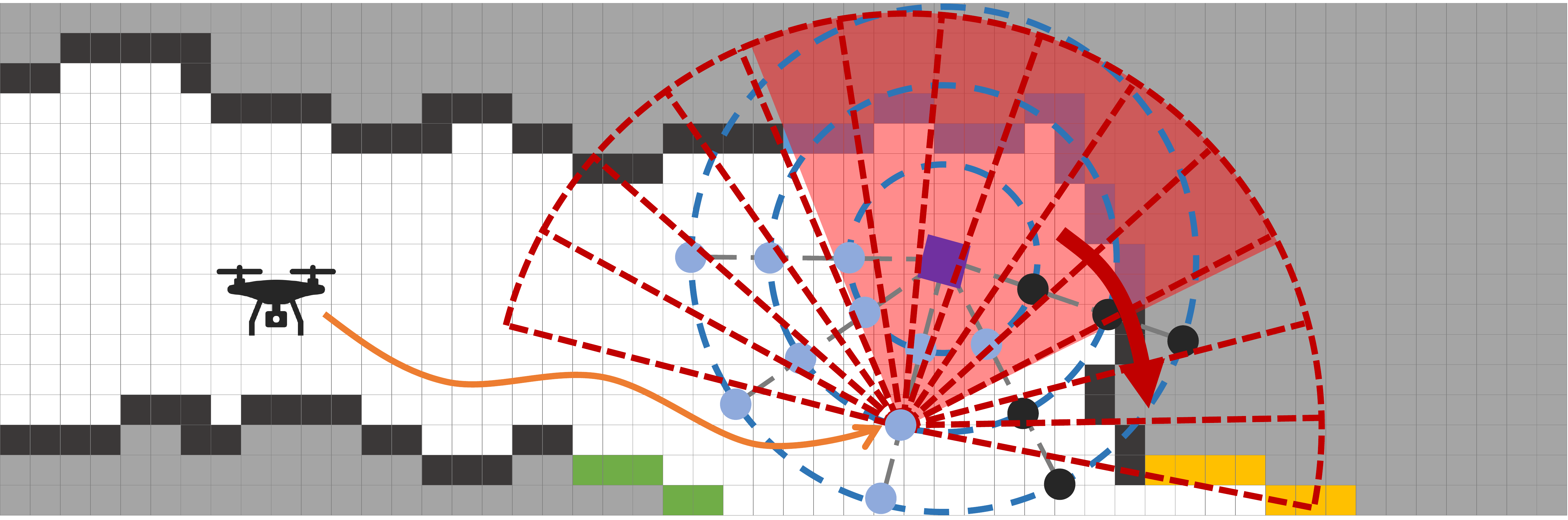}
    \caption{Two-stage viewpoint generation. Feasible Viewpoints (blue dots) are sampled in cylindrical coordinate system from each cluster centroid (purple box). At each viewpoint, predicted information gain is computed in parallel at all sampled yaw angles following a sliding window. }
    \label{fig:vp}
    \vspace{-.3cm}
\end{figure}

\begin{figure}[!t]
    \centering
    \includegraphics[width=1.0\columnwidth]{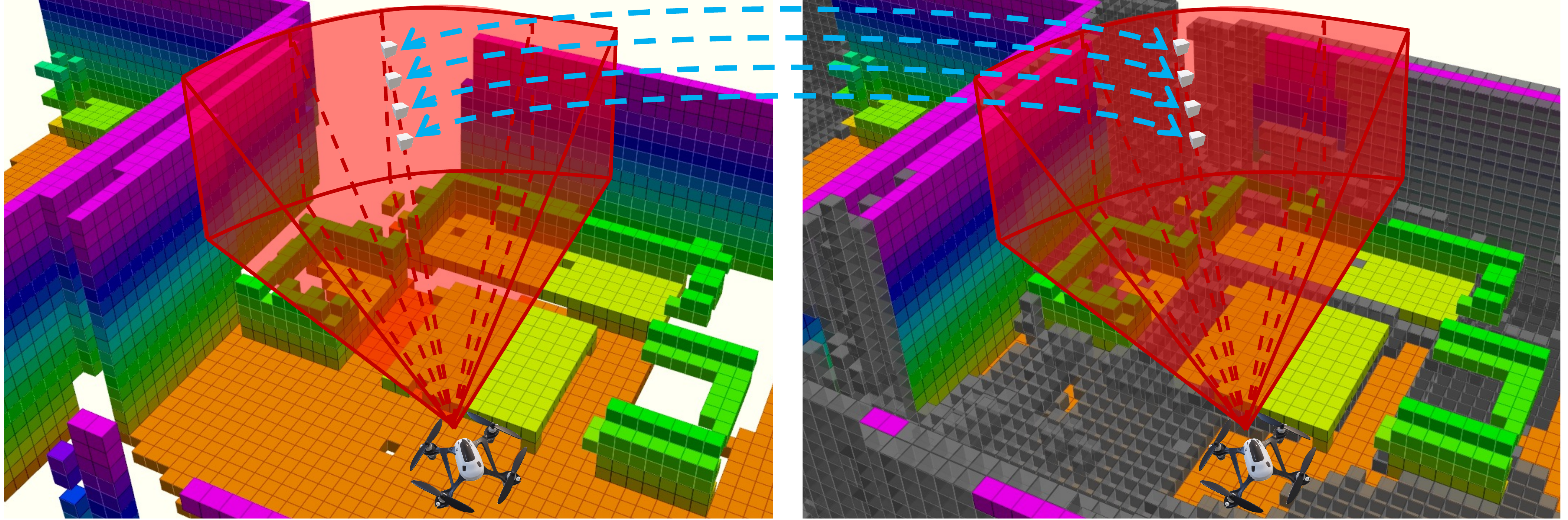}
    \caption{Information prediction process. During ray-casting, `unknown' voxels in the observed map (left panel) are looked up from the map with occupancy prediction (right panel) to predict information gain.}
    \label{fig:info}
    \vspace{-0.5cm}
\end{figure}

The 3-D viewpoint generation and information prediction steps run after the occupancy prediction stage. Inspired by~\cite{zhou2021fuel}, we design a two-stage approach to sample viewpoints $VP_{i}=\{(\mathbf{p}_{i,1}, \xi_{i,1}, g_{i,1}), (\mathbf{p}_{i,2}, \xi_{i,2}, g_{i, 2}), \ldots (\mathbf{p}_{i,n}, \xi_{i,n}, g_{i,n}) \}$ at each $F_{i}$, where $(\mathbf{p}_{i,j}, \xi_{i,j}, g_{i,j})$ represents the 3-D position, yaw angle, and predicted information gain at each sampled viewpoint, respectively. As shown in Fig.~\ref{fig:vp}, in the first stage, the position $\mathbf{p}_{i,j}$ of each viewpoint is uniformly sampled from the cylindrical coordinates that originate from the center of the cluster. At the second stage, centered along the line connects $\mathbf{p}_{i,j}$ to the center of the cluster, multiple yaw angles $\{\xi_{i,j,1}, \xi_{i,j,2}, \ldots \xi_{i,j,m}\}$ are uniformly sampled. Then, we use the methods from~\cite{raptor} to avoid repeated ray-castings and compute in parallel the information gain prediction for all yaw angles sampled. The information prediction process is illustrated in Fig.~\ref{fig:info}. To reduce redundant computations, we calculate the information gain along each slice and aggregate the gains inside each FOV to get the predicted information at different sampled yaw angles. We designed a new information prediction approach by checking voxels from both the current map and the predicted map to better predict the information during ray-casting, as detailed in Algorithm~\ref{alg:info}. After predicting the information gain $g_{i,j,k}$ at each yaw angle $\xi_{i, j, k}$, we save the one with the highest information gain as $\xi_{i,j}$ together with $\mathbf{p}_{i, j}$ and $g_{i,j}$. At the end of the viewpoint generation process, we sort all viewpoints in descending order of predicted information and save the top $N_{vp}$ viewpoints to maximize utility during the planning process. 

\begin{figure}[t]
\vspace{-.4cm}
\begin{algorithm}[H]
\scriptsize
\captionsetup{font=scriptsize} 
\caption{Information Gain Prediction along a ray}\label{alg:info}
\begin{algorithmic} [1]
\State $ray\_start \gets$ ComputeRayStart();
\State $ray\_end \gets$ ComputeRayEnd();
\State $pred\_gain \gets 0$;
\While{! reach ray end}
\State $vox \gets$ next voxel on the ray;
\If{$vox$ ! in map \textbf{or} $vox$ == OCCUPIED}
    \State break;
\EndIf
\If{$vox$ == UNKNOWN}
    \State $in\_pred\_map \gets$ checkCellInPredMap();
    \If{! $in\_pred\_map$}
        \State $pred\_gain ++$;
        \State continue;
    \EndIf
    \State $pred\_vox \gets$ getVoxInPredMap(); 
    \If{$pred\_vox$ == UKNOWN}
    \State $pred\_gain ++$;
    \ElsIf{$predV\_vox$ == OCCUPIED}
    \State $pred\_gain ++$;
    \State break;
    \Else
    \State $pred\_gain ++$;
    \EndIf
\EndIf
\EndWhile
\end{algorithmic}
\end{algorithm}
\vspace{-1cm}
\end{figure}

\subsection{Semantic Object Detection and Frontier Classification}
\label{sec:detection_and_prediction_door}
Indoor environments normally contain rich semantic information which can be used to guide the exploration. In this paper, we study this by detecting doors as semantic objects and classifying frontiers to help the exploration behavioral planning. 
Unlike~\cite{ramon2019semanticIndoor, topo2020indoor}, which detect doors from laser scans, we directly detect doors from occupancy maps around frontier clusters. This allows the detection to happen in parallel at all frontier clusters.

We design a \edited{hand-crafted} two-stage detection algorithm to detect doors on the occupancy map efficiently. At each frontier cluster $F_i$, we extract a small 2D occupancy grid, detect gaps between parallel lines from it using a canny edge detector and probabilistic hough line transform, and compute the position and direction of the gap. In the second stage, we extract the plane perpendicular to the direction of the gap centered at the gap position, which contains the contour of the door. The distance between vertical walls is measured to verify it meets a typical door size. We remove false positive detection by continuously rechecking the detected doors after map updates. All detected doors are initially stored as `to-be-confirmed' until the robot \edited{navigates to \gls{soi} as in Fig.~\ref{fig:bgsm}} to \edited{re-evaluate whether} it's a door.

After semantic object detection, frontier clusters are classified according to the robot's exploration state and detected semantic objects. In this work, we distinguish rooms and corridors and assign the corresponding label to each $F_i$. $F_{i}$ is labeled as `corridor' when the robot is exploring a corridor, navigating towards a door, or exiting a room. While the robot is entering or exploring a room, $F_{i}$ is labeled as `room'. Frontier clusters classified as `room' will be checked to confirm that they are inside the same room as the robot before being considered as a candidate goal during behavioral planning.
    
    \section{Perception-aware planning for exploration}
\label{sec:hybrid_planning}

\subsection{Behavior Planning}
\label{sec:hierarchical_planning_behavior}

\begin{figure}[!t]
    \centering
    \includegraphics[width=1.0\columnwidth]{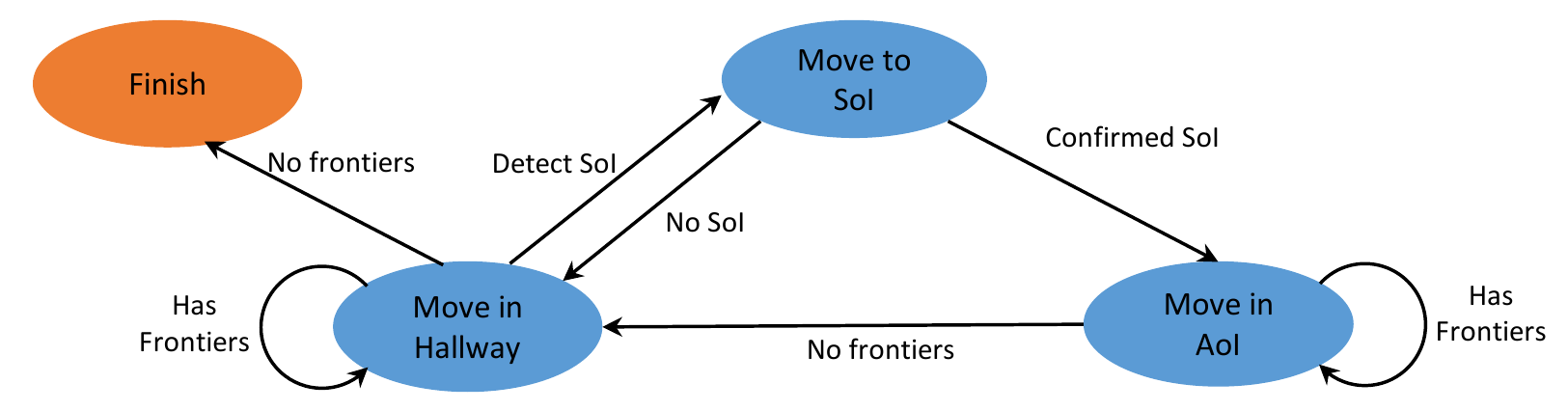}
    \caption{\gls{bgsm}. The \gls{bgsm} takes the abstracted information from the detection and prediction module to make elaborate decisions on next navigation goal during exploration.}
    \vspace{-.5cm}
    \label{fig:bgsm}
\end{figure}

In the structured indoor environment, semantic objects provide strong information about the importance of the region, which classical frontier-based or information-based exploration cannot fully utilize. Assuming that the area of interest should be visited with higher priority and explored thoroughly, we propose a high-level \gls{bgsm}, which selects the appropriate next goal to explore, as illustrated in Fig.~\ref{fig:bgsm}. The navigation behavior is determined by the existence of \glspl{soi}, robot state, and utility of frontiers. For the frontier cluster $F_{i}$, we define the heuristic utility function of viewpoint $vp_{i,j} = (\mathbf{p}_{i,j}, \xi_{i,j}, g_{i,j})$ as follows:
\begin{equation}
    \mathcal{U}(F_{i})  =  \max_{VP_i} u(vp_{i,j}) = \max_{VP_i}  \frac{v_{p,m}}{\text{len}[\mathbf{p}_{e},\mathbf{p}_{i,j}]} g_{i,j}
\end{equation}
where $v_{p,m}$ is the maximum velocity, $\mathbf{p}_{e}$ is the robot's position and \text{len}$[\cdot]$ is the length of a geometric path we search from the current position to the viewpoint. If a robot detects a new \gls{soi} in the corridor, it will navigate to it, confirm it, and enter the \gls{aoi}. Otherwise, it will continuously visit other frontiers according to their computed utilities. Once the robot enter the \gls{aoi}, it will explore the area until no frontiers exist in \gls{aoi}.

\subsection{Predicted Informative Planning}
\label{sec:hierarchical_planning_local}

\subsubsection{Polynomial Trajectory Generations}

Motion planning in 3-D for exploration requires generating flexible, dynamically feasible trajectories. Leveraging the differential flatness of quadrotor dynamics, we optimize the position and yaw trajectories of our agent separately. In particular, we parametrize the trajectory as the output of a 3-D $m$-order integrator, composed of $M$ polynomial segments $\bm{\Phi}(t)$.
Therefore, our trajectory optimization problem becomes
\begin{subequations}
	\begin{align}
	\label{eq:linear}
	& \min_{\Phi(t)} J_c = \int_{t_0}^{t_M} \| \bm{\Phi}^{(\kappa)}(t) \|^2 dt \  \quad  \quad  \quad  \quad \\
	& \ {\rm s.t.} 		
	 \ \mathcal{H}_c(\bm{\Phi}(t), ... ,   \bm{\Phi}^{(\kappa-1)}(t)) =  \bm{0},  \\
 	& \ \quad \quad  \mathcal{G}_d(\bm{\Phi}(t), ... ,   \bm{\Phi}^{(\kappa-1)}(t)) \preceq \bm{0},\quad  \\
       &  \quad \quad \ \forall t \in [t_0, t_M] .
	\end{align}
\end{subequations}
Here $\mathcal{H}_c$ encodes the smoothness, and $\mathcal{G}_d$ represents the dynamic feasibility and obstacle avoidance constraints. We employ the method in~\cite{liu2017ral} to represent the collision-free space as a safe flight corridor. We also leverage the framework in \cite{wang2021generating} to relax the constraints and solve an unconstrained spatial-temporal optimization problem, which is further extended for yaw trajectory optimization.
\subsubsection{Predicted Informative Yaw Planning} 
To generate perception-aware trajectories, we optimize yaw angles with respect to both the predicted information gain and the relative view towards \gls{soi}. Specifically, we first search yaw primitives along the position trajectory with time intervals $\bm{{\rm T}} = [T_1, ..., T_M]^T, T_i = t_i - t_{i-1}$ to get a reference yaw $\psi(t)$ with high-value information. The total cost function to evaluate a primitive segment is defined by:
\begin{equation}
f_{\psi,i}= \beta_{\psi}  \int_{t_{i-1}}^{t_{i}} \ddot{\psi}(t)^2 dt - \frac{T_i}{N_i} \sum_{ \lambda =1}^{N_i} \left ( \beta_{s} f_s(\tilde{t}) + \beta_u g_e(\tilde{t}) \right ), \\\
\end{equation}
where $N_i$ is the sampling number, $\tilde{t}=t_{i-1}+\lambda T_i /N_i$, and $\beta_{\psi}, \beta_{s}, \beta_u $ are non-negative penalties associated with yaw control efforts, perception of semantics, and predicted information gain. The information gain $g_e$ is evaluated using Algorithm~\ref{alg:info}. 
The component of the cost encouraging the \gls{soi} to lie within the horizontal FOV $\theta$ is given by:
\begin{equation}
f_s(t) =  \sigma \left(\frac{\theta}{2} - \left|\psi(t) - {\rm atan2} \left( e_2^T  \bm{p}_r(t), e_1^T  \bm{p}_r(t)\right)\right|\right), 
\end{equation}
where $\sigma(\cdot) = max(\cdot, 0)$, $e_1 = [1, 0, 0]$, $e_2 = [0, 1, 0]$, and $\bm{p}_r(t) = \bm{\Phi}(t) - \bm{p}_s $ is the relative position to the center point $\bm{p}_s$ of the semantic object that the robot navigates to.
After identifying an optimal reference yaw sequence $\gamma= [\gamma_{1}, ..., \gamma_{M-1} ] $ using a search-based procedure, we directly solve for the optimal yaw trajectory $\Psi(t)$ that interpolates the yaw angles  $\bm{ \mathcal{\psi}} = [ \psi_{1}, ..., \psi_{M-1} ]$ at specified times.  The cost function can be expressed as:
\begin{align}
\label{eq:yaw cost}
&J_{\Sigma}(\bm{ \mathcal{\psi}}) = \sum_{i=1}^{M}  \int_{t_{i-1}}^{t_{i}} \Psi^{(3)}(t)^2 dt + \rho_p \sum_{i=1}^{M-1} | \psi_{i}  -\gamma_{i}|^2  + \\
& \rho_v   \sum_{i=1}^{M-1} \sigma  ( | \frac{\psi_{i+1} -  \psi_{i-1}}{T_{i+1} + T_i}  | - v_{\psi, m} )^3 +  \notag \\
& \rho_a \sum_{i=1}^{M-1} \sigma    ( | \frac{(\psi_{i+1} - \psi_{i})/T_{i+1} - (\psi_{i} - \psi_{i-1})/T_{i}}{(T_{i+1} + T_i)/2} | - a_{\psi, m}  )^3,   \notag 
\end{align}
where $ v_{\psi, m}, a_{\psi, m}$ are the maximum yaw velocity and acceleration. $\rho_p, \rho_v,\rho_a $ are constant weights to prevent from undesired and infeasible yaws.

    \section{Results and Analysis}
\label{sec:results}
\begin{table*}[!t]
\caption{EXPLORATION STATISTICS FOR TWO TESTING ENVIRONMENTS}
\scriptsize
\begin{center}
\label{tab:benchmark}
\begin{tabular}{
|P{0.110\textwidth}|
P{0.02\textwidth}P{0.02\textwidth}P{0.02\textwidth}P{0.02\textwidth} | 
P{0.02\textwidth}P{0.02\textwidth}P{0.02\textwidth}P{0.02\textwidth} |   P{0.042\textwidth} |
P{0.02\textwidth}P{0.02\textwidth}P{0.02\textwidth}P{0.02\textwidth} | 
P{0.02\textwidth}P{0.02\textwidth}P{0.02\textwidth}P{0.02\textwidth} |   P{0.042\textwidth} |
}
\hline
& \multicolumn{9}{c|}{\textbf{Small Office}} & \multicolumn{9}{c|}{\textbf{Big Office}} \\ \cline{2-19} 
\textbf{Method} & \multicolumn{4}{c|}{\textbf{Exploration Time (s)}} & \multicolumn{4}{c|}{\textbf{Path Length (m)}} & \multirow{2}{*}{\textbf{Succ(\%)}}  &\multicolumn{4}{c|}{\textbf{Exploration Time (s)}} & \multicolumn{4}{c|}{\textbf{Path Length (m)}} & \multirow{2}{*}{\textbf{Succ(\%)}}  \\ \cline{2-9}  \cline{11-18} 
                                               &   Min & Max & Avg & Std & Min & Max & Avg & Std & &  Min & Max & Avg & Std & Min & Max & Avg & Std & \\ \cline{1-19}
Frontier\cite{Frontier} &  69.79 & 102.2 & 79.88 &  10.61    & 36.52 & 51.69  & 42.90 & 5.47 & 30\% 
& 165.4 & 209.1 & 181.1 & 17.13 & 96.31 & 118.9 & 106.4 & 8.09 & 20\% \\
FUEL\cite{zhou2021fuel}  &  54.68 & 97.65 & 66.18 & 10.88 & 40.21 & 68.28 & 46.76 & 6.51 & \textbf{85\%}
& 150.5 & \textbf{165.4} & 160.8 & \textbf{4.46} & 111.2 & 134.1 & 119.7 & \textbf{6.29} & 40\% \\
Frontier + Util$^{\mathrm{a}}$      & \textbf{51.76} & 95.95 & 72.47 &  12.86  & 34.57 & 65.35  & 47.51 & 8.26 & 50\%
& 134.0 & 196.5 & 156.7 & 21.07 & 93.02 & 145.2 & 108.1 & 18.99 & 25\% \\ 
Frontier + Pred$^{\mathrm{b}}$  & 53.06 & \textbf{78.91} & \textbf{62.92} &  \textbf{7.96}  & 37.34 & 52.72  & 42.70 & \textbf{4.42} & 65\%
& \textbf{128.1} & 178.7 & \textbf{148.8} & 16.03 & 86.31 & 129.6 & 106.4 & 14.60 & 35\% \\
SEER & 58.90
 & 96.52
 & 75.33
 & 11.17
 & \textbf{25.96}
 & \textbf{46.07}
 & \textbf{38.61}
 & 6.06
 & 80\%
& 129.8 & 202.1 & 163.2 & 19.21 & \textbf{76.6} & \textbf{113.4} & \textbf{90.99} & 11.73 & \textbf{75\%}\\ \hline
\multicolumn{18}{p{0.8\textwidth}}{$^{\mathrm{a}}$ Frontier with highest utility, trajectory generation method same as FUEL.} \\
\multicolumn{18}{p{0.8\textwidth}}{$^{\mathrm{b}}$ Frontier with highest utility computed with information prediction, trajectory generation method same as FUEL.}
\end{tabular}
\vspace{-.5cm}
\end{center}
\end{table*}

\subsection{Implementation details}
To achieve an efficient map update, we used a lightweight volumetric mapping module adapted from~\cite{FIESTA}. We implemented OPNet in PyTorch and trained it on the generated dataset. We used an Adam optimizer with an initial learning rate of $10^{-3}$, decays by $0.1$ every 8 epochs, and a batch size of 10. We set $\omega_{occ} = 2$, $\omega_{struct} = 1, \alpha = 5$ and $\beta = 2$. It takes 2.87 hours to train 20 epochs on a machine with an NVIDIA RTX A5000 GPU.

\subsection{Simulation Experiments}

\subsubsection{Information Gain Prediction}
To validate the effectiveness of our proposed information gain prediction module, we first compute the ground truth information gain $g^{gt}_{i, j}$ by comparing the occupancy between the observed map and the ground truth map during exploration at every $VP_i$ sampled from every $F_i$. Then, we compute the information gain with both the proposed method $g_{i, j}$ and the classic method $g^{cls}_{i,j}$ at the same viewpoint $vp_{i,j}$. We compare the mean percentage error between $g^{gt}_{i, j}$ and the information gain estimated by different methods from the $\textbf{22,276}$ data groups. The mean percentage error between $g^{gt}_{i, j}$ and $g^{cls}_{i,j}$ is $\textbf{206.637\%}$ while between $g^{gt}_{i, j}$ and $g_{i,j}$ $\textbf{124.641\%}$. The results show that our method outperforms the classical method by $\approx \textbf{40}\%$.

\subsubsection{Benchmark Results}
\begin{figure}[!t]
    \centering
    \includegraphics[width=1.0\columnwidth]{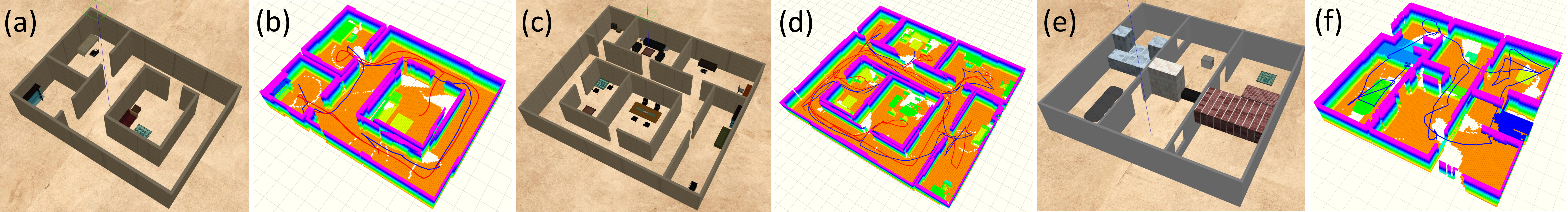}
    \caption{Simulated 3-D Environment in Gazebo and 3-D maps after exploration. Small office (a) of volume $175m^3$. Big Office (c) of volume $392m^3$. Furniture models from \cite{gazebo_models}. Executed trajectories from the proposed method (blue) and FUEL (red) are visualized in the final maps (b, d). 3-D exploration results (f) of the proposed method is demonstrated in the indoor cluttered 3-D environment (e)}.
    \label{fig:sim_env}
    \vspace{-.5cm}
\end{figure}

We have extensively tested our proposed system in Gazebo simulations. Controlled experiments demonstrated how information prediction and perception-aware planning benefit exploration. We benchmarked our proposed method with the state-of-the-art exploration algorithm in both small ($175m^3$) and large ($392m^3$) scale indoor structured environments as shown in Fig,~\ref{fig:sim_env}. The box-bounded maximum velocity is set as 1 $m/s$ and the maximum acceleration is 1 $m/s^2$. The result of running 20 experiments for each method is shown in Tab.~\ref{tab:benchmark}. We record the statistics after a successful exploration, which is defined as having no collisions during the exploration and when the explored volume reaches 95\% of the total space.

 We observed that, by evaluating utilities at the frontiers, the selected candidate goal will lead to more immediate information in general. However, since the information is estimated blindly, it will lead to wrong decisions by selecting a goal that actually provides lower utilities. By utilizing information prediction, information gains at frontiers are estimated much more accurately, which directly improves the quality of selected goals, leading to faster exploration and shorter path lengths. The high-level behavior planning further improves this by reducing repeated visits to the explored region. Instead of maximizing immediate information gain, FUEL minimizes the length of the global tour that covers all frontiers. However, we observed that this coverage tour optimization sometimes leads to repeated visits to the explored regions as shown in Fig.~\ref{fig:sim_env}. This results in a long path length in the structured indoor setting, which can be seen in Tab.~\ref{tab:benchmark}. However, our proposed method uses semantic information to avoid repeated visits to \glspl{aoi} and to plan informative trajectories during exploration. As a result, our method achieves a $\textbf{24\%}$ shorter overall path length with a higher success rate and a comparable exploration time compared with the state-of-the-art method~\cite{zhou2021fuel}. We also observed that our proposed method outperforms other methods in Tab.~\ref{tab:benchmark} more as the environment size increase. 

\subsection{Real World Experiments}

The Falcon 250 UAV used for our experiments is a custom designed platform with a $402\,\text{mm}$ tip-to-tip diameter with $6\,\text{in}$ propellers, weighing $1.29\,\text{kg}$ including a $331\,\text{g}$ 3S $5200$ mAh Li-Po battery. It carries an Intel Realsense camera and an onboard computer as shown in Fig.\ref{fig:fig1}.
\begin{figure}[!t]
    \centering
    \includegraphics[width=1.0\columnwidth]{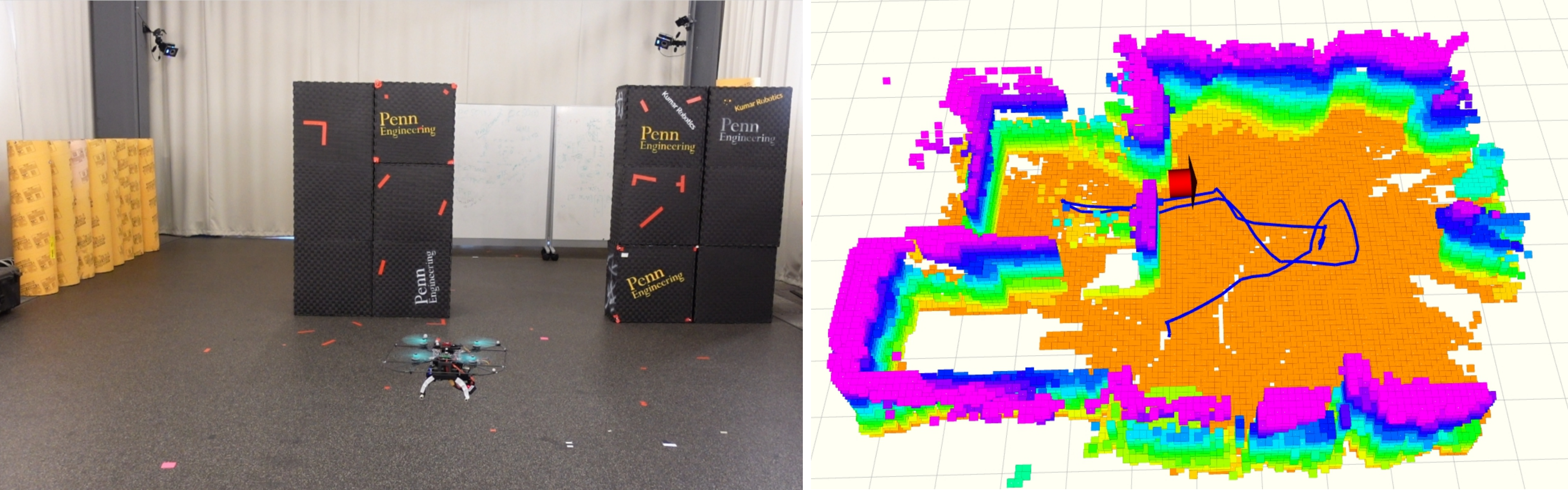}
    \caption{Experiment in a structured 3-D indoor environment. The environment contains a room, a corridor and an open space. We bound the environment with a $12\,m \times 8\,m \times 2.5\,m$ map. The right figure shows the occupancy map after exploration, executed trajectories (blue), and a detected door (red arrow). }
    \label{fig:exp}
    \vspace{-.5cm}
\end{figure}
For real-world experiments, we use the odometry information from the Vicon Motion System. We utilized ONNX to run the trained network model onboard the quadrotor platform. The inference runs at 4 Hz with 4 threads on an i7-10710U CPU. Due to the space limit, we set the box-bounded maximum velocity at $0.4m/s$ and accelerate to $0.5m/s^2$. We conducted extensive real-world experiments in a $12m \times 8m \times 2.5m$ space to validate our proposed framework. As the robot takes off from the corridor and moves forward, it detects a door, navigates towards it, and confirms it. After the door is confirmed, the robot enters the room and completely explores it by navigating to the frontiers that maximize the utilities. The robot then exits the room back to the corridor, selecting the frontier with the highest utility. All utilities are computed considering predicted information gain during this process. The mapping result and the executed trajectories are shown in Figure. \ref{fig:exp}. Real-world experiments prove that our proposed framework is capable of handling realistic structured indoor environments. For more information, we kindly invite the reader to take a look at the video material at \url{https://youtu.be/5ZBkJmCKywg}.

    \section{Conclusion}
\label{sec:conclusion}
In this paper, we develop a novel 3-D exploration framework for autonomous indoor exploration with \glspl{mav}. In particular, we propose an incremental detection and prediction module and a perception-aware planning module to enable \edited{safe} and fast indoor exploration. Our proposed framework shows advantages in total trajectory length and success rate compared with classical and state-of-the-art methods. We validate our proposed framework in real-world experiments in which our proposed framework runs fully onboard a \gls{swap} constrained platform.

\bibliography{literature}

\end{document}